%% file: bmvc_review.tex
\documentclass{bmvc2k}
\usepackage{wrapfig}
\usepackage{xcolor}
\usepackage{tabularx}
\usepackage{amsmath}
\usepackage{graphicx}
\usepackage{hyperref}
\usepackage{mathtools}
\usepackage{wrapfig}
\usepackage{xcolor}
\usepackage{tabularx}
\usepackage{amsmath}
\usepackage{graphicx}
\usepackage{hyperref}
\usepackage{mathtools}
\usepackage{graphicx}
\usepackage{subcaption}
\usepackage{booktabs}
\usepackage{amssymb}
\usepackage{multirow}
\usepackage{tikz}
\usetikzlibrary{fit, backgrounds}
\usetikzlibrary{shapes.geometric, arrows.meta, positioning, calc}
\usepackage{amssymb}\usepackage{pifont}\newcommand{\cmark}{\ding{51}}\newcommand{\xmark}{\ding{55}}\newcommand{\R}{\mathbb{R}}
\tikzset{
    block/.style={rectangle, draw, fill=blue!10, text width=5em, text centered, rounded corners, minimum height=2.5em, thick},
    wideblock/.style={rectangle, draw, fill=blue!15, text width=8em, text centered, rounded corners, minimum height=2.8em, thick},
    smallblock/.style={rectangle, draw, fill=green!15, text width=3.5em, text centered, minimum height=2em, thick},
    lstm/.style={rectangle, draw, fill=purple!15, text width=4em, text centered, rounded corners=2pt, minimum height=2.2em, thick},
    featureblock/.style={rectangle, draw, fill=orange!10, text width=3em, text centered, minimum height=2em, thick},
    arrow/.style={thick,-{Stealth[length=3mm]}},
    dblarrow/.style={thick,{Stealth[length=2.5mm]}-{Stealth[length=2.5mm]}},
    recurrent/.style={thick,-{Stealth[length=2.5mm]}, rounded corners},
}

\usepackage{fontawesome}
\usepackage{lipsum} 

\usepackage[table]{xcolor}
\definecolor{low}{HTML}{FEE8C8}       
\definecolor{moderate}{HTML}{FFF7BC}  
\definecolor{strong}{HTML}{D9F0A3}    
\title{Unapologetically Distributed:\\A Call for Decentralized Document Analysis}

\addauthor{Adrià Molina}{amolina@cvc.uab.cat}{1,2}
\addauthor{Oriol Ramos Terrades}{oriolrt@cvc.uab.cat}{1,2}
\addauthor{Josep Lladós}{josep@cvc.uab.cat}{1,2}

\addinstitution{
 Centre de Visió per Computador\\
 Universitat Autònoma de Barcelona\\
 Bellaterra, Catalonia
}
\addinstitution{
 Computer Science Department\\
 Universitat Autònoma de Barcelona,\\
 Bellaterra, Catalonia
}

\runninghead{A. Molina \etal}{Unapologetically Distributed}

\def\etal{\emph{et al}\bmvaOneDot}

\begin{document}

\maketitle

\begin{abstract}
Privacy has become an increasingly important concern in the Document Analysis community, to the extent that in many environments such as archives, governmental institutions, and local businesses, the adoption of automation is restricted by legal and policy constraints. While federated learning has often been regarded as a ``necessary evil'',  implying an unavoidable performance trade-off in exchange for decentralization and privacy, many prior works overlook its potential to improve robustness to out-of-distribution data. In this paper, we present Unapologetically Distributed, the first comprehensive study evaluating distributed learning in Document Analysis along three key axes simultaneously: the tasks addressed, the architectures employed, and the fine-tuning strategies applied. Specifically, we demonstrate how various distributed training approaches enhance generalization capabilities across diverse tasks such as Table Recognition, handwriting recognition, and Word Spotting, particularly during transfer learning stages. Our results provide strong evidence that decentralization is not merely a constraint, but a valuable opportunity to improve model robustness and adaptability in real-world Document Analysis scenarios.
\end{abstract}


\section{Introduction}

In many Document Analysis scenarios, the assumptions underlying centralized machine learning pipelines are difficult to satisfy. Training modern deep models typically relies on large-scale data aggregation and expensive computational infrastructures, a strategy that has raised growing concerns regarding data privacy and information leakage \cite{carlini2023extracting}. These issues are particularly critical for document data, which frequently contains sensitive, personal, legal, or historical information. As a result, centralization is often impractical or undesirable, despite the availability of valuable training material. This limitation is especially evident in contexts involving privately owned administrative records. Additionally, cultural heritage institutions have long emphasized the importance of preserving documents close to their place of origin, as this practice safeguards their contextual integrity \cite{brothman2001past}, authenticity, and cultural significance \cite{ketelaar2005sharing}, while also aligning with distributed preservation strategies \cite{olliff2021distributed}. Similarly, small and medium-sized enterprises routinely handle confidential documents but lack the resources or legal capacity to externalize data labeling or participate in centralized training schemes. In all cases, the inability to pool data significantly limits the applicability of conventional learning approaches.

\begin{figure}[t]
    \centering
    \begin{tabular}{cc}
\resizebox{.47\textwidth}{!}{\input{tikzpictures/abstract}}
 &  \includegraphics[width=.47\linewidth]{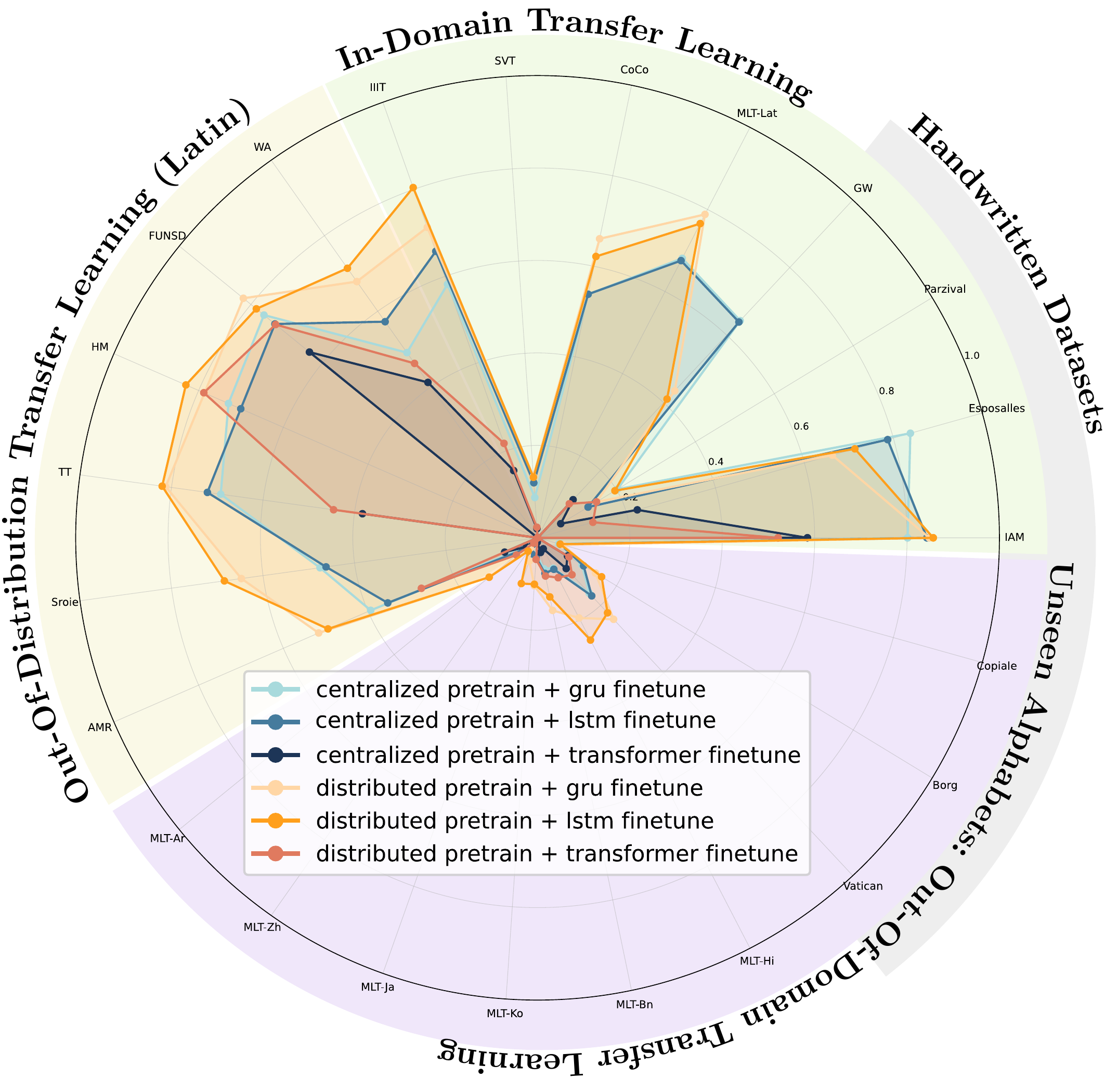} \\
(a) Example of a distributed learning set-up & (b) Word spotting results comparison.
    \end{tabular}
    \caption{We show that distributed regimes (orange area, proposed) consistently outperform centralized pretraining (blue area) across a wide range of architectures and downstream tasks. We perform an extensive evaluation on 27   well-established Word Spotting and recognition datasets.}
    \label{fig:results_radial}
\end{figure}


Distributed learning provides a natural alternative by enabling collaborative model training without direct data sharing, more precisely in our case through post-hoc model merging, where independently trained models are aggregated at the parameter level. However, their adoption in Document Analysis has been hindered by concerns regarding robustness under non-identically distributed data, a setting in which performance degradation has been widely reported \cite{matsuda2022empirical}. At the same time, insights from personalized and meta-learning suggest that distributed training can remain effective when local heterogeneity is explicitly accounted for \cite{arivazhagan2019federated}. This work builds on these observations and investigates distributed learning as a practical and effective solution for Document Analysis in privacy-constrained and low-resource environments.

Our research hypothesis is threefold: 
(i) following model merging principles, features learned by distributed models should be more generalizable and thus serve as better teachers in knowledge distillation; 
(ii) these features should be sufficiently general to enable multi-script learning, even when data distributions differ significantly from the original training data; and 
(iii) the resulting models should provide superior initializations for subsequent in-distribution finetuning stages.

The contribution of this work is to present a large-scale empirical and analytical study of decentralized training in Document Analysis, examining when and why simple model merging strategies yield consistent benefits (see Figure~\ref{fig:results_radial}). Our analysis is structured along three complementary axes: architectural families, task types, and data regimes. We evaluate distributed pre-training across a diverse set of representative tasks which, when considered jointly, cover a broad and practically relevant spectrum of use cases in Document Analysis. By controlling computational budgets, we isolate the effect of decentralized training and characterize the conditions under which it provides measurable advantages. Our objective is not to advocate decentralization as a universal solution, but to characterize the regimes in which it is most effective. In particular, we identify the conditions under which decentralized learning delivers competitive or superior performance; most notably in low-resource, distribution-shifted, and operationally constrained scenarios that frequently arise in real-world Document Analysis applications. To this end, we conduct an extensive evaluation comprising:

\begin{itemize}

    \item \textbf{Cross-modal knowledge distillation}, where features learned in a distributed manner serve as teachers for Word-Spotting students, allowing models to be trained from scratch.
    \item \textbf{Multi-script learning} for new and unseen alphabets in Handwritten Text Recognition tasks, implemented via layer-wise finetuning in a personalized manner.
    \item \textbf{End-to-end, finetuning and zero-shot} evaluations of Table Recognition models initialized through distributed training of Graph Neural Networks.
\end{itemize}

To the best of our knowledge, this article represents the first comprehensive evaluation of distributed learning in Document Analysis that simultaneously considers multiple dimensions: the specific tasks being addressed, the underlying neural architectures employed, and the strategies used for fine-tuning. By systematically examining these aspects, we provide a holistic understanding of how distributedly learned models perform across diverse Document Analysis scenarios and highlight the conditions under which distributed learning provides tangible benefits over centralized approaches.

\section{Related Work}

Existing studies on distributed~\footnote{We note model soups, task arithmetic, model-agnostic meta-learning, and federated learning as distributed learning set-ups, as in all instances different models can be trained distributedly and aggregated \textit{a posteriori}.} document understanding predominantly frame distributed learning as a proxy for centralized performance, rather than as a first-class paradigm capable of reshaping how sensitive tasks on structured documents are learned through multiple domains. 

In \cite{zhang2020fedocr}, a representative example of distributed OCR, the authors note that “\textit{Expectantly, our FedOCR achieves comparable results, which are very close to the results of the centralized training manner},” a formulation that frames federated learning as an approximation of centralized training rather than a standalone paradigm. In their seminal contributions such as \cite{he2021fedgraphnn}, where distributed learning is applied to Graph Neural Network architectures, the conclusions are drawn in the same direction.
In recent years, some works have started to point out that distributedly learned systems may hold significant potential for subsequent fine-tuning strategies. This is the case in \cite{nguyen2024federated}, a pilot study on federated Document Visual Question Answering, where the authors report the outperformance of the centralized regime (C=1, K=1 in their setting, where C denotes the client sampling probability and K the number of participating clients) while also observing that their federated approach achieves comparable results with smaller models in later fine-tuning stages. Nevertheless, the broader implications and global potential of federated learning are not explicitly emphasized. To the best of our knowledge, no prior study has addressed the task of federated learning for Table Recognition, a notable gap given the sensitivity of such data in modern industrial settings. 

In summary, this work is contextualized as the first to holistically demonstrate the potential of distributed learning in Document Analysis across multiple tasks and architectures, while also being the first to perform distributed learning for Table Recognition and knowledge distillation from federatedly learned Document Analysis Systems.
In this paper, readers may find similarities to Model-Agnostic Meta-Learning~\cite{jiang2019improving}, Model Soups~\cite{wortsman2022model}, Model Editing with Task Arithmetic~\cite{ilharco2023editing}, and Federated Learning through \texttt{FedAvg}~\cite{mcmahan2023communication}, depending on their background. We argue that all of the methods above can be treated similarly with regard to their capacity for distributed learning, in the sense that models trained across different computing units can be aggregated without sharing data. 

In contrast to FedAvg, we do not conduct multiple federated rounds. In contrast to MAML, our dataset splits are domain-specific, emulating different institutions. Perhaps the closest related approaches are model editing and model soups; however, neither focuses on the ability of distributedly learned models to incorporate personalized, out-of-domain data. Furthermore, to the best of our knowledge, no prior model soup work has conducted such an extensive evaluation across different architectures, datasets, and application domains.

We note again that all of the aforementioned algorithms share the common principle described in Eq.~\ref{eq:merging}, but implemented within different optimization loops and with emphasis on different aspects of the learning outcomes.

\section{Methodology}

In this study, we aim to evaluate distributed learning holistically across Document Analysis tasks. Specifically, we conduct experiments under three training strategies: knowledge distillation, layer-wise fine-tuning, and end-to-end fine-tuning. This design allows us to assess whether distributedly learned models exhibit higher generality than their centralized counterparts across different Document Analysis scenarios. To support this claim, it is first necessary to clearly define the three methodological settings considered in this work.

Given the broad audience targeted by this study, we begin by introducing the fundamentals of our model merging strategy. Following the notation introduced in \cite{ilharco2023editing}, we present the principles of model merging in Section~\ref{sec:model_mergin}. Readers already familiar with the FedAvg algorithm and the concept of task vectors may directly refer to Section~\ref{sec:meth}, where we detail the specific methodological choices of this study, namely the three fine-tuning strategies evaluated.

\subsection{Model Merging Strategy}
\label{sec:model_mergin}
We consider a \textit{model} as a vector of weights, $\theta = (w_1, w_2, ..., w_m)$ such that $\theta \in \R^m$. Given a model with initial parameters $\theta^{0}$, we train it on a dataset $D$ composed of multiple small sub-sets $d_i$:
\begin{equation}
D = \{d_1, d_2, \ldots, d_n\},
\end{equation}
which converges to a model $\theta^D$. This is what we denote as a \textit{centralized model}, i.e., a set of parameters that have been trained on the dataset $D$ as a whole.

Analogously, a sub-domain $d_i \in D$ optimizes a model that converges to $\theta^{d_i}$. By optimizing on each sub-dataset independently, we obtain different models:

\begin{equation}
\{\theta^{d_1}, \theta^{d_2}, \ldots, \theta^{d_n}\}.
\end{equation}

The parameters of these models (which share a common architecture) can then be averaged as follows:

\begin{equation}
\overline{\theta_{D}} = \frac{1}{n}\sum_{i=1}^{n} \theta^{d_i}.
\label{eq:merging}
\end{equation}

This process is often referred to as task arithmetic \cite{ilharco2023editing}, and it also constitutes the core procedure of the FedAvg algorithm \cite{mcmahan2023communication}. We refer to $\overline{\theta_{D}}$ as a \textit{distributed model}, as it enables each dataset to remain on different computing units while only sharing model weights, but not the underlying data.

\subsection{Finetuning Strategy}
\label{sec:meth}
In this section, three methodological approaches are proposed to verify the hypothesis that \emph{distributedly trained models exhibit superior learning capabilities during the fine-tuning stage}. Specifically, we investigate: (i) knowledge distillation from a frozen encoder in a Word Spotting task; (ii) layer-wise fine-tuning strategies for handwritten text recognition (HTR) and optical character recognition (OCR) problems when adapting models to new alphabets; and (iii) graph-based Table Recognition using end-to-end fine-tuned models. Through these approaches, we aim to present a broad and transversal methodology that encompasses a wide range of use cases, architectures, and training strategies, thereby providing a robust and empirical validation of the proposed hypothesis.

\paragraph{\textbf{Knowledge Distillation in Word Spotting}}
Knowledge distillation is the training strategy consisting in transfering information from a base network (teacher), which is usually trained with high computational and data resources, to a smaller (student) network.
For this task, we aim to leverage the information of a frozen vision encoder into a small text encoder to perform a Query-by-String evaluation of a Word Spotting system. According to our hypothesis, distributed models should behave better as teachers due to its higher generalization capability.

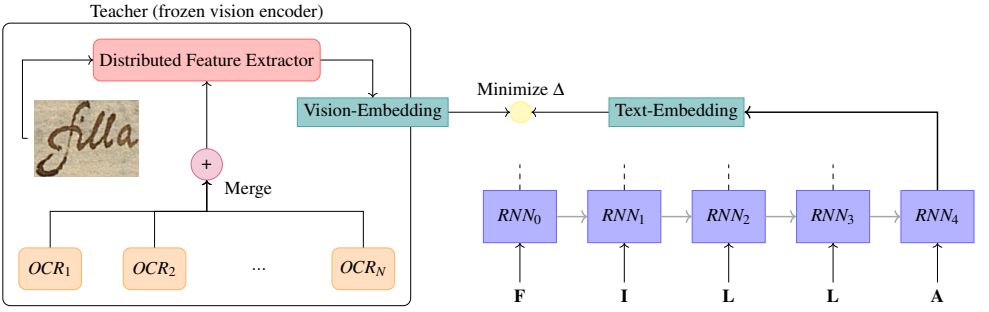
\begin{figure}[t]
    \centering
    \resizebox{\textwidth}{!}{\input{tikzpictures/kdistillation}}
    \caption{Knowledge distillation scheme: multiple pretrained OCRs are combined into a frozen teacher, which guides an RNN to map text to images in unseen alphabets or domains.}
    \label{fig:knowledge_distillation}
\end{figure}

As seen in Figure~\ref{fig:knowledge_distillation}, different off-the-shelf OCR models from~\cite{rodriguez2025ocr} are sampled as Vision Encoders dedicated to both scene and handwritten text recognition. After removing the language layer, we perform model merging (see Equation~\ref{eq:merging}) to obtain a distributedly learnt feature extractor (i.e. an extractor for which the different training datasets are not shared). Then, given a target dataset (the one we want to perform distillation on), visual embeddings are computed.
For the textual embeddings, we make use of standard bidirectional Recurrent Neural Networks (LSTM \cite{hochreiter1997long} and GRU \cite{chung2014empirical}) and a transformer encoder architecture \cite{atienza2021vision}. In all the cases, the text encoder is computed by adding a special \texttt{<RET>} token in the last position. The distance between textual and vision embeddings are minimized through a Triplet Margin Loss \cite{balntas2016learning}, transferring in consequence the spatial distribution of the visual embeddings into the text encoder.

\paragraph{\textbf{Multi-Script Learning in Text Recognition}}

\begin{figure}[t]
    \centering
    \resizebox{\textwidth}{!}{\input{tikzpictures/ocr}}
    \caption{Distributed models trained on Latin alphabets are evaluated on unseen scripts by averaging backbones and fine-tuning only the language head.}
    \label{fig:ocr}
\end{figure}
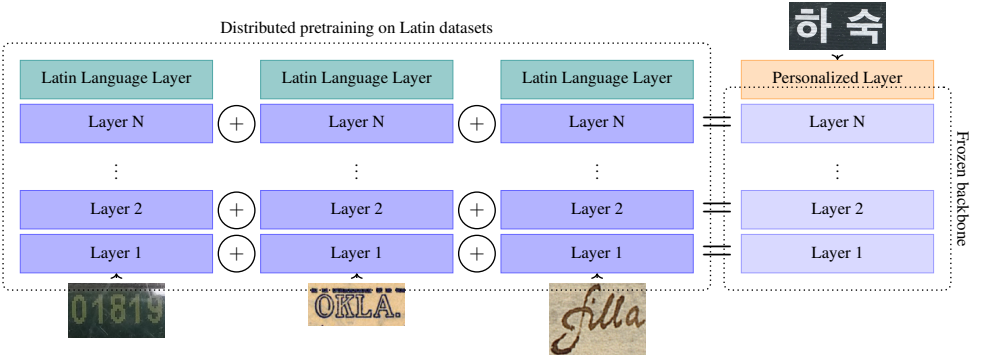

In a personalized learning framework, the goal is for a federated model to effectively incorporate new data, typically through layer-wise fine-tuning. In this section, we adopt a standard Vision Transformer (ViT) architecture as our reference model (see \cite{atienza2021vision}), pre-trained on the HierText dataset \cite{long2022towards}. This model serves as our \textit{baseline}.

Starting from this baseline, we conduct pre-training using a collection of datasets described in Section~\ref{sec:datasets}. Two training paradigms are considered: (i) a \textit{centralized} setting, where all datasets are combined and used to train a single model, and (ii) a \textit{distributed} setting, where separate models are trained on individual datasets and later merged through model merging.

Since all pre-training datasets predominantly contain Latin characters (along with arabic numerals), we frame the recognition of an entirely new alphabet as a personalized learning task. Specifically, we perform personalized fine-tuning on ciphered and multilingual datasets (see Section~\ref{sec:datasets}), using limited data and a small number of epochs, and restricting updates to the later stages of the network. As shown in Figure \ref{fig:ocr}, to accommodate the new alphabet, the OCR language modeling head is replaced with a single layer whose output dimension matches the size of the new character dictionary.

Under our hypothesis, the distributed pre-training approach should enable the model to incorporate personalized data more effectively, resulting in improved recognition performance compared to the centralized model, when trained under identical hyperparameter settings and number of epochs.

Additionally, we report results for experiments that do not rely on HierText pre-training. In this alternative setup, models are initialized randomly (using the same seed for both centralized and distributed approaches), trained independently on each dataset, and subsequently merged via model merging in the distributed case.

\paragraph{\textbf{Fine-tuning in Graph-Based Table Recognition}}

We also include experiments regarding End-to-End training of Graph Neural Networks in the context of Table Recognition; where, as seen in \cite{riba2022table}, we consider text regions in a document as nodes consisting of their $(x_1, y_1, x_2, y_2)$ coordinates and a \textit{vector class} consisting of a histogram of the alphanumeric characters of the OCR text as node features. Then, a visibility graph is constructed with edges connecting adjacent nodes. The objective of the message passing algorithm is to classify each node as table, non-table or header categories only from its geometry, connectivity and the histogram of characters. For further details on the graph construction we refer readers to the cited original implementation.

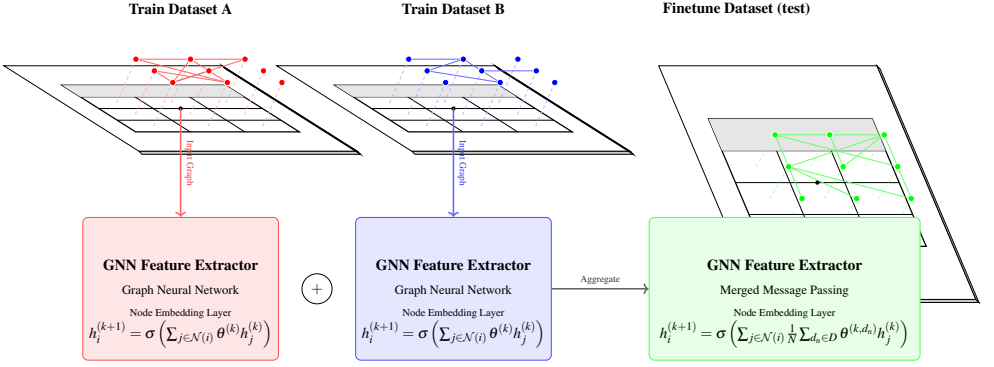
\begin{figure}
    \centering
    \resizebox{\textwidth}{!}{\input{tikzpictures/table}}
    \caption{Distributed learning of message passing layers. Standard models (red/blue) are combined via distributed message passing (green), aggregating edge parameters ($\theta$) while keeping node features ($h_j$) fixed.}
    \label{fig:table_abs}
\end{figure}

As seen in Figure~\ref{fig:table_abs}, we train $N$ graph-based Table Recognition models independently (hence, distributedly) and then merge their message passing layers (i.e. the weights of a Graph Neural Network). We, then, fine-tune the model on an unseen dataset ($N+1$) and test the performance in this fine-tuning dataset. Namely, each i-th node is updated on a k-th step with weights $\theta^K$ as follows:

\begin{equation}
        h_i^{(k+1)} = \sigma\left(\sum_{j \in \mathcal{N}(i)} \theta^{(k)} h_j^{(k)}\right)
        \label{eq:normal_message_passing}
\end{equation}

Where $h^k_i$ denotes the node $i$ features during the message passing step $k$, $\sigma$ represents a non-linearity and $j \in \mathcal{N}$ denotes the set of nodes connected to $i$. Consequently, for a distributed message passing model the following expression is used:

\begin{equation}
        h_i^{(k+1)} = \sigma\left(\sum_{j \in \mathcal{N}(i)} \frac{1}{N}\sum_{d_n \in D} \theta^{(k, d_n)} h_j^{(k)}\right)\label{eq:distributed_message_passing}
\end{equation}

Where the message is computed by using the k-th layer on every GNN trained on a different distributed dataset $d_n$ and then averaged to obtain a distributedly learned representation. It is important to note that the node representation and the edge construction requires to be the same in both source and target tasks. This works for systems on which node features are agnostic to the training data. 
\begin{table}[h]
    \centering

\begin{tabular}{ccccc}
\toprule
Dataset Name & Tasks & Finetuning & Pretraining & Alphabet \\
\midrule
IAM \cite{marti2002iam} & WS & KD & \cmark & Latin \\
Esposalles \cite{romero2013esposalles} & WS & KD & \cmark & Latin \\
George Washington \cite{fischer2012lexicon} & WS & KD & \cmark & Latin \\
Parzival  \cite{fischer2012lexicon}& WS & KD & \cmark & Latin \\
CoCoText \cite{veit2016cocotext} & WS & KD & \cmark & Latin \\
MLT19 (Latin) \cite{nayef2019icdar2019} & WS & KD & \cmark & Latin \\
SVT \cite{wang2011end} & WS & KD & \xmark & Latin \\
IIIT5K \cite{mishra2012scene} & WS & KD & \xmark & Latin \\
MLT19 (Arabic) \cite{nayef2019icdar2019} & WS, OCR & KD / PL & \xmark & Arabic \\
MLT19 (Chinese) \cite{nayef2019icdar2019} & WS, OCR & KD / PL & \xmark & Chinese \\
MLT19 (Japanese) \cite{nayef2019icdar2019} & WS, OCR & KD / PL & \xmark & Japanese \\
MLT19 (Korean) \cite{nayef2019icdar2019} & WS, OCR & KD / PL & \xmark & Korean \\
MLT19 (Hindi) \cite{nayef2019icdar2019} & WS, OCR & KD / PL & \xmark & Devanagari \\
MLT19 (Bangla) \cite{nayef2019icdar2019} & WS, OCR & KD / PL & \xmark & Bangla \\
Copiale \cite{knightCopialeCipher2011} & WS, HTR & KD / PL & \xmark & Ciphered \\
Borg \cite{aldarrabBorgLat8982018} & WS, HTR & KD / PL & \xmark & Ciphered \\
Vatican \cite{heder2022decode} & WS, HTR & KD / PL & \xmark & Mixed \\
Historical Maps \cite{weinman2019deep} & WS & KD & \xmark & Latin \\
SROIE \cite{karatzas2013icdar} & WS & KD & \xmark & Latin \\
FUNSD \cite{jaume2019funsd} & WS & KD & \xmark & Latin \\
AMR \cite{laroca2019convolutional} & WS & KD & \xmark & Digits \\
TotalText \cite{ch2020total} & WS & KD & \xmark & Latin \\
WordArt \cite{xie2024icdar} & WS & KD & \xmark & Latin \\
ICDAR2019 \cite{gao2019icdar} & TR & E2E & \cmark & Latin \\
RVL-CDIP \cite{harley2015evaluation} & TR & E2E & \cmark & Latin \\
con-anonym \cite{riba2022table} & TR & E2E & \cmark & Latin \\
M96 & TR & E2E & \xmark & Latin \\

\bottomrule
\end{tabular}
\caption{Overview of the 27 evaluation datasets. For each dataset, we indicate whether it is in-pretraining or out-of-domain.}
    \label{tab:datasets}
\end{table}
\section{Experimental Setup}
The employed pre-training and fine-tuning datasets, as well as the evaluation setup, are defined in this section. When referring to a fine-tuning dataset, we denote the process in which a model, after being pre-trained using either a centralized or distributed strategy, is fine-tuned on that dataset individually. The effectiveness of the fine-tuning stage is then assessed using the corresponding test partition of the same dataset.

\subsection{Datasets}
\label{sec:datasets}
In this Section, we list the different 27 employed datasets and the rationale behind the choice of each of them. For a comprehensive summary of the datasets and tasks used for this study reader may refer to Table \ref{tab:datasets}.

\paragraph{\textbf{Knowledge Distillation in Word Spotting}}
For the Word Spotting task, we first train an encoder which, as introduced in the methodology section, serves as a frozen teacher within a knowledge distillation framework. This encoder is trained using what we define as \emph{in-domain} HTR and OCR datasets, which share a common Latin alphabet and consist of well-cropped, centered text samples with limited font variability. 

Specifically, the handwritten text datasets include IAM \cite{marti2002iam}, Esposalles \cite{romero2013esposalles}, George Washington and Parzival \cite{fischer2012lexicon}, while the scene text datasets comprise CoCoText \cite{veit2016cocotext} and the Latin split of MLT19 \cite{nayef2019icdar2019}. Due to their limited contribution to the overall training corpus (stemming from their reduced size and narrow scope) the SVT \cite{wang2011end} and IIIT5K \cite{mishra2012scene} datasets are excluded from the training process and instead used as control variables to evaluate performance on unseen yet still in-domain datasets.

Subsequently, knowledge from the trained vision encoder is distilled into a text encoder for the Query-by-String Word Spotting task. We define the Historical Maps \cite{weinman2019deep}, SROIE \cite{karatzas2013icdar}, and FUNSD \cite{jaume2019funsd} datasets as \emph{out-of-domain} due to the presence of printed artifacts acting as background distractors in word images. Similarly, the WordArt dataset \cite{xie2024icdar} and TotalText \cite{ch2020total} are considered out-of-domain because of their inclusion of artistic and highly variable fonts. The AMR dataset \cite{laroca2019convolutional} is also employed as a fine-tuning dataset, motivated by the presence of glares and visual artifacts that partially occlude textual information. In addition to the aforementioned out-of-domain datasets containing primarily Latin characters, we further incorporate the Arabic, Chinese, Japanese, Korean, Hindi, and Bangla partitions of MLT19 \cite{nayef2019icdar2019} as both \emph{out-of-domain} and \emph{out-of-vocabulary} fine-tuning datasets, representing the most challenging scenario for knowledge distillation. Finally, the Copiale \cite{knightCopialeCipher2011}, Borg \cite{aldarrabBorgLat8982018}, and Vatican cipher datasets \cite{heder2022decode} are included under the same rationale, as they introduce previously unseen symbol systems and vocabularies. We report the performance of this fine-tuning strategy across all previously described datasets, including in-domain, out-of-domain Latin, and out-of-domain out-of-vocabulary scenarios. For each dataset, a different model and independent is fine-tuned for ten epochs  and evaluated on the corresponding test partition.
\paragraph{\textbf{Multi-Script Learning in Text Recognition}}

In the case of multi-script personalized Text Recognition, we are only interested in the capacity of the model to integrate highly personalized data, hence, data that significantly differs from the original training distribution. We make use of the same latin alphabets datasets as in the previous section. Namely, we train a latin OCR model using IAM \cite{marti2002iam}, Esposalles \cite{romero2013esposalles}, George Washington \cite{fischer2012lexicon}, Parzival, CoCoText \cite{veit2016cocotext} and the Latin split of MLT19 \cite{nayef2019icdar2019}.

Because of Latin Out-of-Distribution data constitutes nothing but an extension of the original training data, we only consider non-latin alphabets as suitable for a personalized learning evaluation. We include fine-tuning on the Arabic, Chinese, Japanese, Korean, Hindi, and Bangla partitions of MLT19 \cite{nayef2019icdar2019}and the Copiale \cite{knightCopialeCipher2011}, Borg \cite{aldarrabBorgLat8982018}, and Vatican cipher datasets \cite{heder2022decode}. We include both the performance for personalized OCR systems for every language and personalized approaches to multi-linguality (i.e. learning all new alphabets simultaneously).

\paragraph{\textbf{Fine-tuning in Graph-Based Table Recognition}}
Because of the sparsity in categories of Table Recognition benchmarks, we have taken the decision to simplify every dataset to contain only Table~/~Not-Table labels, hence posing a binary segmentation problem.
We have conducted experiments with the well-known dataset from the ICDAR 2019 Competition \cite{gao2019icdar}, RLV-CDIP \cite{harley2015evaluation} and Con-Anonym~\cite{riba2022table}. Additionally, and perhaps more interestingly, we present an evaluation on a privately owned dataset (M96\footnote{An anonymized version will be produced for reproducibility.}) consisting of only 50 page-level annotations. Hence yielding insights on the capacity of such models to integrate new distributions of data in a low-resource set-up.

\subsection{Metrics and evaluation}
The evaluation protocol on the presented experiments is constructed as follows:
First, for every task, a model is trained with every data available as training dataset; this is a centralized model containing all the training datasets information, noted as $\theta^D$. For every individual dataset, we also train its corresponding individual model with the same number of steps as the centralized model $\theta^{d_n}$. These models are aggregated following Equation \ref{eq:merging}, which yields the distributed version of $\theta^D$, $\overline{\theta_{D}}$ 

Note that training $N$ data points ($N=n_1+n_2$) during $T$ epochs / steps causes $N \times T$ forward/backward steps. In the case of decomposing the dataset, the total cost is $n_1 \times T + n_2 \times T$ forward / backward steps. Note that product is distributive, therefore this expression is converted to $T \times (n_1 + n_2)$, which knowing that $n_1$ and $n_2$ are parts of N ($N=n_1+n_2$), the total cost and the data seen by each model is exactly equivalent.

Second, we consider a fine-tuning dataset, for which we train (using its train partition) during 25 epochs every model (distributed or centralized). This specialized model is tested on its fine-tuning task using the corresponding metric for each one of the showcased applications; this is: 
\begin{figure}[t]
    \centering
    \includegraphics[width=0.99\linewidth]{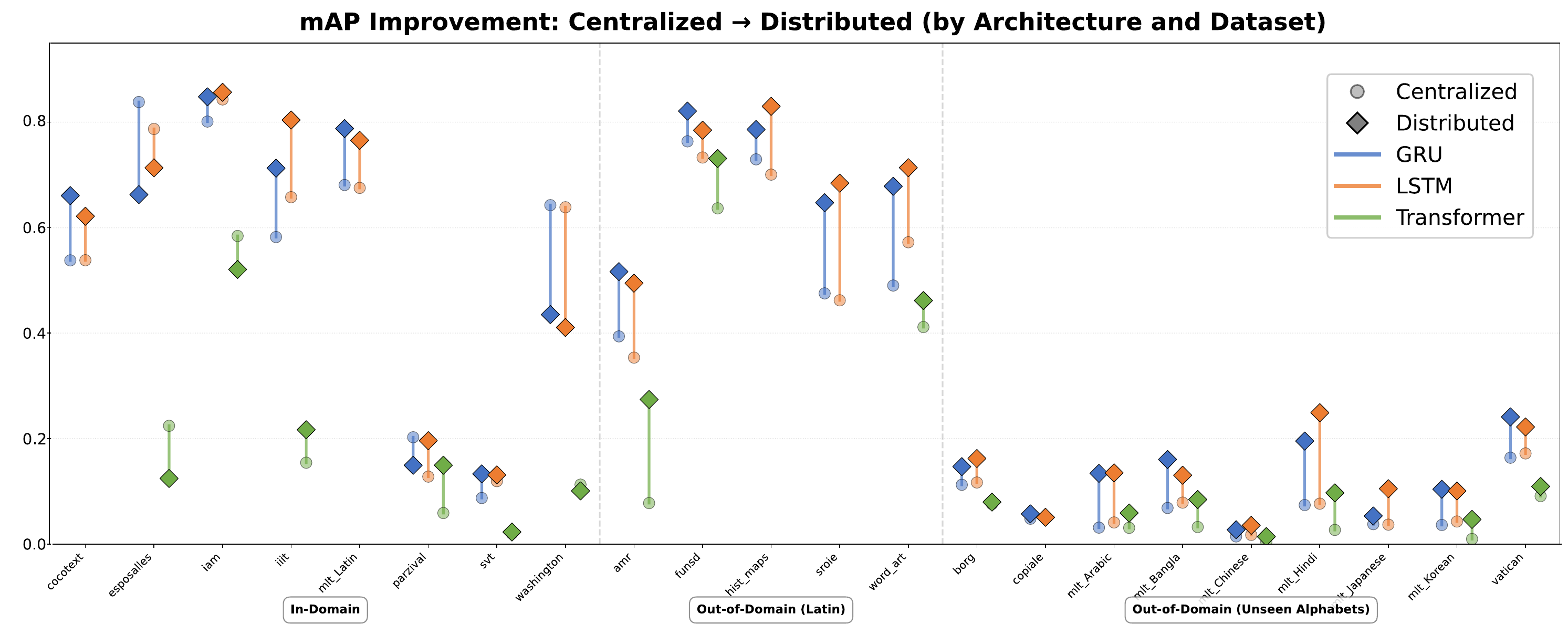}
    \caption{Mean Average Precision (mAP) for Query-by-String Word Spotting across 22 datasets using GRU, LSTM, and Transformer. Distributed (squares) excels in out-of-domain tasks, especially with shared alphabets.}
    \label{fig:results_whole}
\end{figure}

\paragraph{\textbf{Word Spotting}}
We follow a Query-by-String (QbS) evaluation protocol using the mean Average Precision (mAP) metric. In this approach, given a set of test queries (text strings) and a gallery of test images (word crops), the model retrieves images by minimizing the cosine distance between the query text embedding and the image embeddings.

The mean Average Precision (mAP) is computed as the average of the Average Precision (AP), defined as:
\begin{equation}
\text{AP}(q) = \frac{1}{N_q} \sum_{k=1}^{n} P(k)\cdot \text{rel}(k),
\end{equation}
where \(N_q\) is the number of relevant (ground-truth) images for that query and \(n\) is the total number of retrieved images. Here, \(P(k)\) represents the precision at rank \(k\), defined as the number of relevant images in the top \(k\) results divided by \(k\), and \(\text{rel}(k)\) is a binary relevance indicator that equals 1 if the image at rank \(k\) is relevant to the query and 0 otherwise.
\paragraph{\textbf{Text Recognition}}

We evaluate character recognition performance using word-level accuracy. We conduct experiments under two training paradigms: (1) language-specific models trained independently on each language, and (2) a single multi-lingual model trained jointly on all languages to assess cross-lingual transfer capabilities.

\textbf{Word-level Accuracy} measures the percentage of words that are transcribed perfectly without any character errors:

\begin{equation}
\text{Accuracy} = \frac{\text{\# correctly transcribed words}}{\text{\# total words}} \times 100\%
\end{equation}




\paragraph{\textbf{Table Recognition}}

We evaluate Table Recognition performance using a graph-based approach with two complementary metrics: node accuracy and edge accuracy. The model represents document structure as a graph where nodes correspond to semantic regions and edges encode spatial relationships between them. In one hand, \textbf{node accuracy} reflects how well the model identifies the type of each region. In our case, nodes represent regions labeled as either table or non-table. A node is considered correct if the predicted label matches the ground truth. On the other, \textbf{edge accuracy} measures how well the model captures spatial relationships between regions. This includes relationships within the same region, between regions of the same type, and between regions of different types. An edge is correct if both its presence and relationship type match the ground truth.
\input{data/bigtable/table_ws_summary}








\section{Results}
This section presents results for the three tasks introduced earlier. We analyze when distributed learning setups improve fine-tuning with knowledge distillation for Word Spotting, personalized text recognition, and E2E table detection.
\subsection{Word Spotting}
The main results of knowledge distillation for Query-by-String Word Spotting are shown in Figure~\ref{fig:results_whole}. For latin datasets included in pre-training, results are mixed: fine-tuning from distributed setups occasionally degrades performance, yet six out of eight datasets benefit from distributed pre-training in at least one architecture, yielding improvements in 75\% of cases. Out-of-domain fine-tuning consistently favors distributed pre-trained models. Datasets not seen during centralized or distributed pre-training show clear gains, including for unseen alphabets. Performance improvements are larger when the fine-tuning alphabet matches that of pre-training, highlighting both the generality and limitations of the learned features. Overall, Table~\ref{tab:comparison} reports an average improvement of 5\% in mean average precision. In-domain gains can be negative, but out-of-domain datasets show substantially larger improvements across both Latin and unseen alphabets. Transformer-based text encoders achieve the lowest absolute performance, likely due to higher data requirements, yet relative gains from distributed pre-training are comparable to LSTM encoders, with an average increase of 6\% in mean average precision.

\subsection{Optical Text Recognition}

Table~\ref{tab:results_lang} shows results for three training setups: (1) fine-tuning from a baseline trained only on HierText, (2) centralized pre-training on HierText and a multilingual dataset before per-language fine-tuning, and (3) the fully distributed version. We also include distributed learning from random initialization without auxiliary pre-training. We observe a consistent advantage of our personalized learning approach with respect the centralized and baseline approaches, with an average improvement of $\times1.62$ in the best-case scenario. In some cases, such as the Borg cipher, distributed learning is essential for the weights to accommodate the inclusion of such a different domain. Needless to say, as it was specified on previous sections, both centralized and distributed pretraining have performed the same amount of forward-backward steps on the same exact data; hence, we can only attribute the incapacity of the centralized approach to incorporate certain languages to the training regime itself.  Additionally, we include results for multilingual training (Table~\ref{tab:combined_acc}). In contrast to the previous experiments, where languages were fine-tuned independently, these settings consider an optical character recognition system trained to learn all alphabets simultaneously. Under this formulation, we observe that the performance gap between centralized and distributed training regimes becomes narrower. Notably, accuracy for the Borg cipher is fully recovered. This suggests that distributed models adapt more effectively in low-resource scenarios, whereas centralized approaches are sufficient when abundant data is available in the personalization step. Nevertheless, even in these settings, the distributed approach outperforms the centralized model.

\begin{table}[t]

\resizebox{\textwidth}{!}{
\begin{tabular}{l|ccc|cccccc|c}
\toprule
 & Vatican & Borg & Copiale & Arabic & Chinese & Japanese & Korean & Bangla & Hindi & $\times\Delta$ \\
\midrule
From Baseline & .549 & .382 & .825 & .131 & .020 & .116 & .282 & .260 & .470 & - \\
Centr. (base) & .465 & .000 & .794 & .175 & .0103 & .0951 & .214 & .115 & .384 & 0.69 \\
Dist. (base) & \textbf{.591} & \textbf{.505} & .838 & \textbf{.422} & \textbf{.073} & \textbf{.197} & \textbf{.374} & \textbf{.462} & \textbf{.514} & \textbf{1.62} \\
Dist. (random) & .480 & .272 & \textbf{.930} & .410 & .010 & .114 & .271 & .266 & .282 & 1.04 \\
\bottomrule
  \end{tabular}}
\caption{Transfer learning accuracy across languages. Distributed fine-tuning from a shared pretrained model ($Z$) outperforms centralized fine-tuning and task arithmetic from scratch, especially on low-resource alphabets.}
\label{tab:results_lang}

\end{table}

\begin{table*}[t]
\centering
\small
\caption{Multi-lingual and multi-cipher training results.}
\label{tab:combined_acc}

\setlength{\tabcolsep}{5pt}
\renewcommand{\arraystretch}{1.15}

\begin{tabular}{lccccccccc}
\toprule

& \multicolumn{6}{c}{\textbf{Multi-Lingual}} 
& \multicolumn{3}{c}{\textbf{Multi-Cipher}} \\

\cmidrule(lr){2-7}
\cmidrule(lr){8-10}

\textbf{Accuracy $\uparrow$}
& Arabic
& Bangla
& Chinese
& Hindi
& Japanese
& Korean
& Borg
& Copiale
& Vatican \\

\midrule

Distributed
& \textbf{.472}
& \textbf{.469}
& \textbf{.127}
& \textbf{.539}
& \textbf{.252}
& \textbf{.435}
& \textbf{.573}
& \textbf{.840}
& \textbf{.566} \\

Centralized
& .402
& .385
& .076
& .468
& .183
& .324
& \textbf{.573}
& .819
& .524 \\

\bottomrule
\end{tabular}
\end{table*}
\subsection{Table Detection}
The results for graph-based recognition, reported in Table~\ref{tab:table_recog}, correspond to a leave-one-out evaluation at the dataset level. In this setup, models are trained on all datasets except the one indicated by the corresponding column. For instance, zero-shot results in the RLV column are obtained by training on M96, ICDAR, and Con-Anonym while excluding RLV from pre-training. Conversely, end-to-end results in the ICDAR column correspond to models pre-trained on M96, RLV, and Con-Anonym and subsequently fine-tuned on ICDAR. Zero-shot results, where no fine-tuning is performed on the target dataset, consistently favor the distributed approach in both node and edge accuracy. With partial fine-tuning of the upper layers, performance becomes mixed, while end-to-end fine-tuning generally benefits the centralized model, with the exception of M96. These results support our hypothesis: distributed models are most effective in low-resource regimes, either when no adaptation is possible (zero-shot) or when only limited data is available (e.g., M96). As data availability and adaptation capacity increase, performance progressively shifts toward centralized training, with partial fine-tuning representing an intermediate regime.

\begin{table}[]
\resizebox{\textwidth}{!}{
\begin{tabular}{@{}l|cc|cc|cc|cc@{}}
\cmidrule(l){2-9}
                         & \multicolumn{2}{l|}{M96}                                   & \multicolumn{2}{l|}{RLV}                                          & \multicolumn{2}{l|}{ICDAR}                                        & \multicolumn{2}{l}{CON-Anonym}                                   \\ \cmidrule(l){2-9} 
                         & \multicolumn{1}{l}{Node Acc} & \multicolumn{1}{l|}{Edge Accuracy} & \multicolumn{1}{l}{Node Acc} & \multicolumn{1}{l|}{Edge Accuracy} & \multicolumn{1}{l}{Node Acc} & \multicolumn{1}{l|}{Edge Accuracy} & \multicolumn{1}{l}{Node Acc} & \multicolumn{1}{l}{Edge Accuracy} \\ \midrule
Centr. (zero-shot)  & 87.50\%                      & 85.70\%                            & 76.70\%                      & 72.90\%                            & 68.10\%                      & 78.00\%                            & \textbf{75.50\%}             & 70.10\%                           \\
Dist. (zero-shot)  & \textbf{88.00\%}             & \textbf{87.00\%}                   & \textbf{78.10\%}             & \textbf{78.00\%}                   & \textbf{79.00\%}             & \textbf{84.70\%}                   & 73.00\%                      & \textbf{77.20\%}                  \\ \midrule
Centr. (partial)    & 87.60\%                      & 85.20\%                            & 76.90\%                      & 69.40\%                            & 82.60\%                      & \textbf{81.70\%}                   & \textbf{89.30\%}             & 73.80\%                           \\
Dist. (partial)    & \textbf{88.00\%}             & \textbf{86.50\%}                   & \textbf{79.70\%}             & \textbf{72.30\%}                   & \textbf{83.40\%}             & 78.70\%                            & 89.10\%                      & \textbf{75.40\%}                  \\ \midrule
Centr. (E2E) & 86.10\%                      & 69.90\%                            & 85.30\%                      & 81.50\%                   & 92.00\%       & \textbf{95.50\%}                   & \textbf{97.80\%}             & \textbf{90.60\%}                  \\
Centr. (\cite{riba2022table}) & -                      & -                            & 67.06\%                      &    \textbf{ 83.02\% }              &  \textbf{ 92.20\%}          & 91.60\%                 &     84.74\%        &  89.42\%           \\

Dist. (E2E) & \textbf{88.90\%}             & \textbf{78.30\%}                   & \textbf{86.10\%}             & 79.80\%                            & 90.60\%                      & 94.60\%                            & 97.70\%                      & 89.80\%                           \\ \bottomrule
\end{tabular}}
\caption{Table Recognition results: Node (cell) and Edge (link) accuracy. Models are trained on three datasets and fine-tuned on the fourth.}
\label{tab:table_recog}
\end{table}
\section{Conclusions}
In this study, we conducted an extensive set of comparative experiments between centralized and distributed learning paradigms. Our analysis goes beyond the commonly cited privacy and security benefits of distributed learning, and instead focuses on identifying the conditions under which it also provides performance advantages. This naturally raises the question: under which circumstances does distributed learning yield the greatest gains? To achieve sound conclusions in the study, we have designed a comprehensive set of experiments corresponding to a variety of settings in terms of tasks, learning scenarios and datasets. In the Word Spotting experiments, where only the text encoder is trainable and the vision encoder remains frozen, results (Figure~\ref{fig:results_whole}) are mixed when fine-tuning is performed in-domain. However, in out-of-domain settings, distributed approaches consistently dominate. This effect is particularly pronounced for GRU- and LSTM-based text encoders (Table~\ref{tab:comparison}) which, in contrast to Transformer-based architectures, achieve strong performance with substantially lower data requirements. For character recognition, distributed learning exhibits its largest advantage when languages are learned independently (Table~\ref{tab:results_lang}). When fine-tuning data is abundant, as in the multilingual setting (Table~\ref{tab:combined_acc}), the performance gap narrows. In graph-based Table Recognition, contrary to our initial expectations, end-to-end fine-tuning proves to be the least favorable scenario for distributed models (Table~\ref{tab:table_recog}).

\begin{table}[t]
\caption{Summary of the conclusions and general performance trends. }
\label{tab:summ_conts}
\centering
\small

\begin{tabular}{lcccccc}
\toprule
Regime & Low-Data & ZS & OOD & Param.-Efficient & E2E & ID \\
\midrule
Distributed 
& \cellcolor{strong} Strong
& \cellcolor{strong} Strong
& \cellcolor{strong} Strong
& \cellcolor{strong} Strong
& \cellcolor{moderate} Moderate 
& \cellcolor{low} Low\\

Centralized
& \cellcolor{moderate} Moderate
& \cellcolor{low} Low
& \cellcolor{moderate} Moderate
& \cellcolor{moderate} Moderate
& \cellcolor{strong} Strong &\cellcolor{strong} Strong \\
\bottomrule
\end{tabular}

\end{table}
What overarching pattern emerges from these findings? From an architectural perspective, distributed learning performs best in scenarios with a limited number of learnable parameters. In Table Recognition, performance is strongest in the zero-shot setting, becomes mixed when only part of the model is fine-tuned, and degrades under full end-to-end fine-tuning. Similarly, in Word Spotting, lightweight architectures such as GRUs and LSTMs benefit more from distributed learning than heavier Transformer models. These results suggest that practitioners employing parameter-efficient architectures or operating under constraints that prevent fine-tuning can particularly benefit from distributed pre-training. From a data-centric perspective, distributed learning is most effective in low-resource regimes. In Table Recognition, the largest gains are observed on the smallest dataset, M96. A similar trend appears in character recognition, where personalization to new alphabets yields a larger advantage when languages are trained independently than in joint multilingual training, where data availability is higher and the performance gap diminishes. Overall, these findings (summarized in Table~\ref{tab:summ_conts}) indicate that distributed pre-training is especially advantageous for practitioners working with limited data or in low-resource scenarios.

In conclusion, the question raised by the title becomes clear: decentralization in Document Analysis is not a universal prescription, but a targeted call. It is addressed to practitioners operating under low-resource conditions, limited adaptability, or stringent deployment constraints, for whom distributed learning consistently delivers competitive and often superior performance. Domains such as historical Document Analysis stand out as particularly well aligned with this paradigm. In these settings, annotations are rarely abundant, and data distributions often diverge significantly from those of modern, large-scale datasets. Decentralized learning offers a promising pathway in such contexts, further reinforced by the privacy and access constraints commonly imposed by cultural heritage institutions and archives. A similar alignment can be found in industrial environments involving privately owned administrative records, where automation is hindered by the sensitivity of the data and the infeasibility of large-scale annotation. In this regard, small and medium-sized enterprises, which frequently handle confidential documents without the resources or legal capacity to externalize a labeling procedure, represent a natural and highly relevant application domain for distributed Document Analysis.

\clearpage
\section*{Acknowledgments}
\footnotesize{
This work has been partially supported by the Spanish project PID2024-157778OB-I00, Ministerio de Ciencia e Innovación, the Departament de Cultura of the Generalitat de Catalunya, and the CERCA Program. Adrià Molina is funded with the PRE2022-101575 grant provided by MCIN / AEI / 10.13039 / 501100011033 and by ERDF/EU. We extend our gratitude to Jialuo Chen for his work on adapting \cite{riba2022table} to our framework.
}

{\small\bibliography{egbib}}

\clearpage
\appendix
\section*{Supplementary Material}
This text correponds to the supplementary material for the BMVC2026 paper \textbf{\textit{Unapologetically Distributed: A Call for Decentralized Document Analysis }}. The notes found in the following sections answer some of the questions arised during the review stage, and may be useful for a part of the audience.

\subsection*{Single-round merging as distributed learning}
We choose single-round aggregation because it is the unique operation shared by FedAvg, task arithmetic, model soups and 1-epoch meta-learning (Eq.~3): it lets readers from the federated, model-merging and meta-learning communities read the paper jointly, which is our stated goal. Advanced federated optimizers (FedProx, SCAFFOLD) are corrections to client drift \emph{across rounds}; at one round they collapse to FedAvg, so under our protocol they introduce no distinct comparison. Regarding multi-round FedAvg, our internal ablation on Table Recognition (Table~\ref{tab:rounds}) shows rounds 2--5 shift node/edge accuracy by at most 1--2 points, with no monotonic trend and no change to which regime wins.

\subsection*{Statistical significance}
Table 2 aggregates 22 datasets, so the reported gains act as paired comparisons across many tasks rather than a single run. A one-sided Wilcoxon signed-rank test over all paired centralized/distributed Word Spotting runs confirms significance:
\[
p = 8.8\times10^{-6}\;(n{=}70\text{ pairs; dist.\ wins }57).
\]
Table~\ref{tab:rounds} additionally shows small spreads across five independent aggregation repetitions. 

\subsection*{$\times$1.62 and claim strength}
All rows of Table 3, including Centr.(base), share the same HierText pretraining stage; only Dist.(random) starts from scratch, and it still reaches $\times$1.04, which we consider evidence of robustness rather than a hidden weakness. We will label $\times$1.62 explicitly as best-case in the text. The negative in-domain gains (Table 2) are deliberate findings: the paper does not claim universal superiority, and Table 6 exists precisely to delimit where distributed learning helps (low-resource, zero-shot, OOD, parameter-efficient) and where it does not (in-domain E2E). We will soften any remaining absolute wording.

\subsection*{Why does merging generalize better?}
As an intuition, parameter averaging acts as an implicit smoothing operator over the loss landscape. Each independently trained model $\theta^{d_i}$ converges to a minimum shaped in part by dataset-specific noise, directions that fit peculiarities of $d_i$ but carry no transferable signal. Since these idiosyncratic directions are largely uncorrelated across the $n$ independent runs, averaging them cancels them out, while directions that are consistently reinforced across datasets, corresponding to genuinely shared, task-relevant structure, survive and dominate the resulting $\overline{\theta}_D$. The merged model effectively behaves as a coarse ensemble collapsed into a single set of weights, inheriting the flatter, wider regions of the loss surface that independent minima tend to share, which is also the usual explanation for why flat minima generalize better. This is consistent with the pattern observed throughout the paper: distributed pre-training helps most when the downstream task or architecture leaves little room to re-specialize away from this shared component (low-resource, zero-shot, parameter-efficient regimes), and helps least when full end-to-end fine-tuning can freely override it.
A full theoretical treatment would require an article of its own; we want to note the experimental basis is comprehensive, in consequence, we position this paper as the empirical foundation such theory needs.

\begin{table}[h]
\resizebox{\columnwidth}{!}{
\setlength{\tabcolsep}{10pt}
\begin{tabular}{@{}l|cc|cc|cc|cc@{}}
\toprule
& \multicolumn{2}{c|}{M96} & \multicolumn{2}{c|}{RLV} & \multicolumn{2}{c|}{ICDAR} & \multicolumn{2}{c}{CON} \\
& Node & Edge & Node & Edge & Node & Edge & Node & Edge \\ \midrule
ZS -- r2 & 88.6 & 87.3 & 76.9 & 79.6 & 80.6 & 76.9 & 72.4 & 75.7 \\
ZS -- r3 & 88.1 & 86.7 & 75.0 & 78.6 & 83.5 & 82.2 & 74.5 & 76.7 \\
ZS -- r4 & 88.2 & 86.8 & 75.2 & 79.0 & 82.2 & 78.7 & 72.5 & 79.4 \\
ZS -- r5 & 87.6 & 86.7 & 74.3 & 78.2 & 82.3 & 81.8 & 68.7 & 78.9 \\ \midrule
Part -- r2 & 88.6 & 87.0 & 80.9 & 72.3 & 83.5 & 81.4 & 88.5 & 76.4 \\
Part -- r3 & 88.2 & 86.2 & 78.7 & 71.8 & 83.8 & 81.4 & 89.0 & 76.6 \\
Part -- r4 & 88.2 & 86.4 & 80.0 & 72.5 & 82.5 & 79.6 & 88.6 & 74.8 \\
Part -- r5 & 87.8 & 86.3 & 78.6 & 71.2 & 82.5 & 79.0 & 88.9 & 75.7 \\ \midrule
E2E -- r2 & 88.7 & 77.0 & 86.6 & 82.2 & 91.0 & 94.1 & 97.6 & 91.9 \\
E2E -- r3 & 89.0 & 73.5 & 86.0 & 85.9 & 91.7 & 95.0 & 97.7 & 92.1 \\
E2E -- r4 & 87.6 & 80.0 & 86.5 & 84.0 & 90.3 & 92.6 & 97.7 & 89.8 \\
E2E -- r5 & 88.4 & 80.9 & 86.0 & 85.7 & 91.9 & 95.3 & 97.7 & 91.2 \\ \bottomrule
\end{tabular}}
\caption{Multi-round FedAvg on Table Recognition, rounds 2--5 (round 1 and centralized in submitted Tab.~5). No consistent gain beyond round 1.}
\label{tab:rounds}
\end{table}

\end{document}

%% file: tikzpictures/abstract.tex
\begin{tikzpicture}[
    arrow/.style={->, thick, color=gray!70},
    connector/.style={thick, color=gray!70},
    trainblock/.style={draw, rectangle, minimum width=1.9cm, minimum height=0.9cm,
                       fill=blue!30, draw=blue!60, align=center},
    modelblock/.style={draw, rounded corners, rectangle, minimum width=1.9cm,
                       minimum height=0.9cm, fill=orange!25, draw=orange!60, align=center},
    mergeblock/.style={draw, rounded corners, rectangle, minimum width=1.9cm,
                       minimum height=0.9cm, fill=purple!20, draw=purple!60, align=center},
    transferblock/.style={draw, rectangle, minimum width=1.9cm, minimum height=0.9cm,
                          fill=red!25, draw=red!60, align=center},
    targetblock/.style={draw, rectangle, rounded corners=2pt,
                        minimum width=1.9cm, minimum height=0.9cm,
                        fill=teal!40, draw=teal!70, align=center}
]

\def\offset{0.05}
\def\dataset_separation{2}

\newcommand{\placeDataset}[4]{%

  \coordinate (first) at (4*\offset, #1-4*\offset);
  \coordinate (last)  at (0,          #1);

  \coordinate (dataset_a) at ($(first)!0.5!(last)$);

  \node
    at ($(dataset_a) + (1, -0.7)$) {#4};

  \node[trainblock] (centralized_a)
    at ($(dataset_a) + (2cm,0)$) {Centralized\\Training};

  \node[modelblock, right=0.5cm of centralized_a]
    (pretrained_a) {Pretrained\\Model};

  \draw[arrow] (dataset_a) -- (centralized_a);
  \draw[arrow] (centralized_a) -- (pretrained_a);

  \coordinate (#3) at (pretrained_a.east);

  \foreach \i in {4,3,2,1,0} {%
    \node[opacity=0.9] at (\i*\offset, #1-\i*\offset)
      {\includegraphics[width=1cm]{#2}};
  }
}

\placeDataset{0}{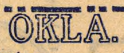}{data_d}{Train with dataset D}
\placeDataset{2}{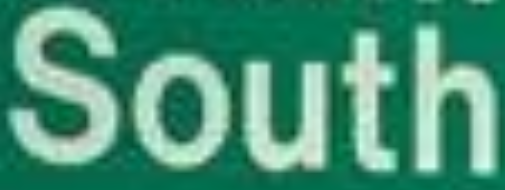}{data_c}{Train with dataset C}
\placeDataset{4}{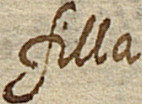}{data_e}{Train with dataset B}
\placeDataset{6}{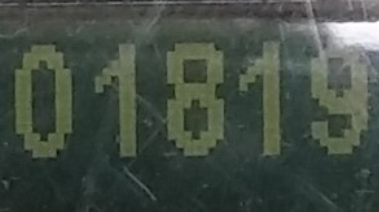}{data_f}{Train with dataset A}

\path
  let
    \p1 = (data_d),
    \p2 = (data_c),
    \p3 = (data_e),
    \p4 = (data_f),
    \n1 = {(\y1 + \y2 + \y3 + \y4)/4}
  in
coordinate (centroid) at (\x1 + 4, \n1);

\draw[connector] (data_d) -| (centroid);
\draw[connector] (data_c) -| (centroid);
\draw[connector] (data_e) -| (centroid);
\draw[connector] (data_f) -| (centroid);

\node[mergeblock, right=0.8cm of data_f] (merge_models)
  {Model\\Merging};

\draw[arrow] (centroid) |- (merge_models.west);

\node[modelblock]
  (pretrained_merged)
  at ($(merge_models.south) + (0,-1)$) {Pretrained\\Model};

\draw[arrow] (merge_models.south) -- (pretrained_merged.north);

\node at ($(data_f.north) + (0,0.8)$) {Merge distributed models};

\node[transferblock]
  (transfer) at ($(pretrained_merged.south) + (0,-2.5)$) {Transfer\\Learning};

\node[targetblock]
  (target) at (transfer |- data_d)
  {Target\\Model};

\draw[arrow] (transfer.south) -- (target.north);
\draw[-, thick, color=gray!70] (pretrained_merged.south) -- (transfer.north);

\node[opacity=0.9] (unseen) at ($(pretrained_merged.south) + (0,-1)$)
  {\includegraphics[width=1cm]{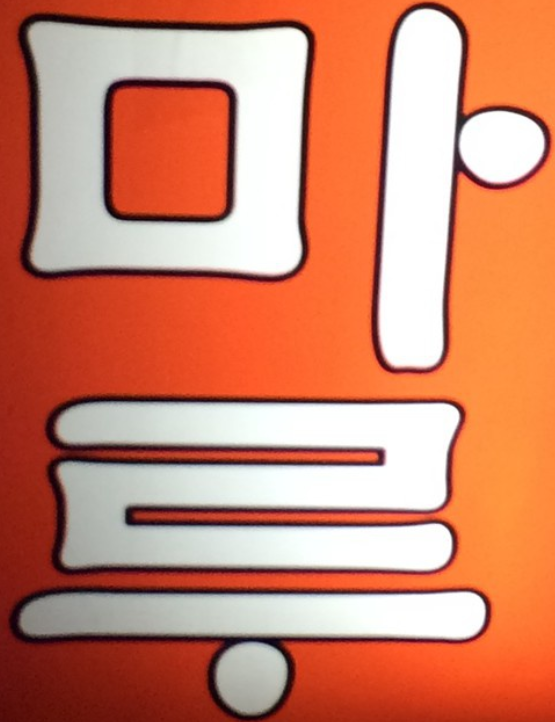}};

\node[align=center, rotate=270, anchor=west, font=\scriptsize]
  at ($(unseen.east) + (0.1, 1)$) {New Alphabet};

\end{tikzpicture}

%% file: tikzpictures/kdistillation.tex
\begin{tikzpicture}[
    rnn/.style={draw, rectangle, minimum width=1.4cm, minimum height=1cm, fill=blue!30, draw=blue!60},
    emb/.style={draw, rectangle, fill=teal!40, draw=teal!70},
    arrow/.style={->, thick, color=gray!70}
]

\def\initial_x_ocr{-12}       
\def\per_ocr_separation{2}    
\def\ocr_y{-1}                
\def\agg_y{1}                 
\def\arrow_top_y{0.1}         

\node[draw, rounded corners, minimum width=1.2cm, minimum height=0.8cm, fill=orange!25, draw=orange!60] (ocr1) at (\initial_x_ocr,-1) {$OCR_1$};
\node[draw, rounded corners, minimum width=1.2cm, minimum height=0.8cm, fill=orange!25, draw=orange!60] (ocr2) at ({\initial_x_ocr + \per_ocr_separation},-1) {$OCR_2$};
\node (moreocrs) at ({\initial_x_ocr + 2*\per_ocr_separation},-1) {$...$};
\node[draw, rounded corners, minimum width=1.2cm, minimum height=0.8cm, fill=orange!25, draw=orange!60] (ocrN) at ({\initial_x_ocr + 3*\per_ocr_separation},-1) {$OCR_N$};

\coordinate (agg_center) at ({\initial_x_ocr + 1.5*\per_ocr_separation}, \agg_y);
\node[draw, circle, minimum size=0.6cm, fill=purple!20, draw=purple!60] (agg) at (agg_center) {+};
\node[anchor=north west] at (agg.south east) {Merge};

\coordinate (ocr1_top) at (ocr1.north |- 0,\arrow_top_y);
\coordinate (ocr2_top) at (ocr2.north |- 0,\arrow_top_y);
\coordinate (ocrN_top) at (ocrN.north |- 0,\arrow_top_y);

\draw[->] (ocr1.north) -- (ocr1_top) -| (agg.south);
\draw[->] (ocr2.north) -- (ocr2_top) -| (agg.south);
\draw[->] (ocrN.north) -- (ocrN_top) -| (agg.south);

\node[draw, rounded corners, minimum width=1.2cm, minimum height=0.8cm, fill=red!25, draw=red!60] (merged) at ({\initial_x_ocr + 1.5*\per_ocr_separation}, \agg_y + 2) {$\text{Distributed Feature Extractor}$};
\draw[->] (agg.north) -> (merged.south);
\node (img) at ([xshift=-2cm, yshift=0.5cm]agg.west) {\includegraphics[width=2cm]{images/assets/esposalles_filla.png}};

\coordinate (img_helper) at ([xshift=-0.1cm]img.west);

\draw (img.west) -- (img_helper);
\draw[->] (img_helper) |- (merged.west);

\node[draw, rounded corners, line width=0.5pt, 
      fit=(ocr1.south west) (ocrN.south east) (merged.north), 
      inner xsep=0.3cm, inner ysep=0.3cm] 
      (teacher) {};
\node (text_teacher) at ([yshift=0.2cm]teacher.north) {Teacher (frozen vision encoder)};

\def\initial_x_lstm{-3}       
\def\per_lstm_separation{2}    

\node[rnn] (h1) at (\initial_x_lstm, 0) {$RNN_0$};
\node[rnn] (h2) at (\initial_x_lstm + \per_lstm_separation * 1,0) {$RNN_1$};
\node[rnn] (h3) at (\initial_x_lstm + \per_lstm_separation * 2,0) {$RNN_2$};
\node[rnn] (h4) at (\initial_x_lstm + \per_lstm_separation * 3,0) {$RNN_3$};
\node[rnn] (h5) at (\initial_x_lstm + \per_lstm_separation * 4,0) {$RNN_4$};

\node (x1) at ([yshift=-1cm]h1.south) {\textbf{F}};
\node (x2) at ([yshift=-1cm]h2.south) {\textbf{I}};
\node (x3) at ([yshift=-1cm]h3.south) {\textbf{L}};
\node (x4) at ([yshift=-1cm]h4.south) {\textbf{L}};
\node (x5) at ([yshift=-1cm]h5.south) {\textbf{A}};

\draw[->, thin] (x1) -- (h1);
\draw[->, thin] (x2) -- (h2);
\draw[->, thin] (x3) -- (h3);
\draw[->, thin] (x4) -- (h4);
\draw[->, thin] (x5) -- (h5);

\draw[arrow] (h1) -- (h2);
\draw[arrow] (h2) -- (h3);
\draw[arrow] (h3) -- (h4);
\draw[arrow] (h4) -- (h5);


\draw[dashed] (h1) -- +(0,1);
\draw[dashed] (h2) -- +(0,1);
\draw[dashed] (h3) -- +(0,1);
\draw[dashed] (h4) -- +(0,1);

\node[emb] (emb) at (0,2) {Text-Embedding};
\node[emb] (vision_emb) at ([xshift=1cm]merged.east |- emb) {Vision-Embedding};
\draw[->] (merged.east) -| (vision_emb.north);
\draw[->, thick] (h5) |- (emb);
\coordinate (emb_center) at ($ (emb.east)!0.5!(vision_emb.west) $);
\node[draw, circle, minimum size=0.4cm, fill=yellow!30, draw=yellow!70] (minimize) at (emb_center) {};

\node[anchor=south] at (minimize.north) {Minimize $\Delta$};

\draw[->] (emb.west) -- (minimize.east);
\draw[->] (vision_emb.east) -- (minimize.west);

\end{tikzpicture}

%% file: tikzpictures/ocr.tex
\begin{tikzpicture}
\tikzstyle{layer} = [
  draw,
  minimum width=4cm,
  minimum height=8mm,
  align=center
]
\def\LayerSep{0.1}
\def\LayerDist{0.9}
\def\ModelSep{5}
\newcommand{\PlaceLayers}[5]{%
    \node[layer, fill=blue!30, draw=blue!60] (#3) at (#1,0) {Layer 1};
    \node[layer, fill=blue!30, draw=blue!60, above=\LayerSep of #3] (l2-#3) {Layer 2};
    \node[above=\LayerSep of l2-#3] (dots-#3) {$\vdots$};
    \node[layer, fill=blue!30, draw=blue!60, above=\LayerSep of dots-#3] (ln-#3) {Layer N};
    \node[layer, fill=teal!40, draw=teal!70, above=\LayerSep of ln-#3] (#5) {#4};
    \ifnum\pdfstrcmp{#2}{None}=0
    \else
        \node[below=\LayerSep of #3] (img-#3) {\includegraphics[width=2cm]{#2}};
        \draw[->, thick] (img-#3.north) -- (#3.south);
    \fi
}
\newcommand{\PlaceLayersFrozen}[5]{%
    \node[layer, fill=blue!15, draw=blue!40] (#3) at (#1,0) {Layer 1};
    \node[layer, fill=blue!15, draw=blue!40, above=\LayerSep of #3] (l2-#3) {Layer 2};
    \node[above=\LayerSep of l2-#3] (dots-#3) {$\vdots$};
    \node[layer, fill=blue!15, draw=blue!40, above=\LayerSep of dots-#3] (ln-#3) {Layer N};
    \node[layer, fill=orange!25, draw=orange!60, above=\LayerSep of ln-#3] (#5) {#4};
    \ifnum\pdfstrcmp{#2}{None}=0
    \else
        \node[below=\LayerSep of #3] (img-#3) {\includegraphics[width=2cm]{#2}};
        \draw[->, thick] (img-#3.north) -- (#3.south);
    \fi
}
\pgfmathsetmacro{\xposone}{0}
\pgfmathsetmacro{\xpostwo}{\xposone + \ModelSep}
\pgfmathsetmacro{\xposthree}{\xpostwo + \ModelSep}
\pgfmathsetmacro{\xposfour}{\xposthree + \ModelSep}

\PlaceLayers{\xposone}{images/assets/amr.png}{l1coord}{Latin Language Layer}{langlayer1}
\PlaceLayers{\xpostwo}{images/assets/maps.png}{l1coord2}{Latin Language Layer}{langlayer2}
\PlaceLayers{\xposthree}{images/assets/esposalles_filla.png}{l1coord3}{Latin Language Layer}{langlayer3}
\PlaceLayersFrozen{\xposfour}{None}{l1coord4}{Personalized Layer}{langlayer4}

\foreach \i in {0,1,3} {
    \node[circle, draw=black, thick] at ($(l1coord)!0.5!(l1coord2) + (0,\i*\LayerDist)$) {\large$+$};
    \node[circle, draw=black, thick] at ($(l1coord2)!0.5!(l1coord3) + (0,\i*\LayerDist)$) {\large $+$};
    \node at ($(l1coord3)!0.5!(l1coord4) + (0,\i*\LayerDist)$) {\Huge $=$};
}

\node[draw, dotted, rounded corners, thick, fit={(l1coord) (langlayer3)}, inner sep=10pt, label={[anchor=south]above:Distributed pretraining on Latin datasets}] {};

\coordinate (ghost-below-lang4) at ([yshift=-\LayerSep]langlayer4.south |- ln-l1coord4.north);

\node[above=\LayerSep of langlayer4] (jap-img) {\includegraphics[width=2cm]{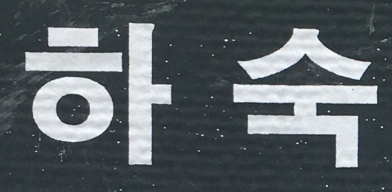}};
\draw[->, thick] (jap-img.south) -- (langlayer4.north);

\node[draw, dotted, rounded corners, thick, fit={(l1coord4) (ghost-below-lang4)}, inner sep=10pt, label={[anchor=south, rotate=-90]right:Frozen backbone}] {};
\end{tikzpicture}

%% file: tikzpictures/table.tex
\begin{tikzpicture}
\def\rotangle{25}  

\newcommand{\calculateprojection}[3]{
    \path let \p1 = (#1) in
        \pgfextra{
            \pgfmathsetmacro{\origx}{\x1/1cm}
            \pgfmathsetmacro{\origy}{\y1/1cm}
            \pgfmathsetmacro{\newx}{\origx*cos(\rotangle) - \origy*sin(\rotangle)*sin(#3)}
            \pgfmathsetmacro{\newy}{\origy*cos(#3)}
            \coordinate (#2) at (\newx, \newy);
        };
}

\newcommand{\calculateprojectionz}[4]{
    \path let \p1 = (#1) in
        \pgfextra{
            \pgfmathsetmacro{\origx}{\x1/1cm}
            \pgfmathsetmacro{\origy}{\y1/1cm}
            \pgfmathsetmacro{\zoffset}{#2}
            \pgfmathsetmacro{\newx}{\origx*cos(\rotangle) - \origy*sin(\rotangle)*sin(#4) + \zoffset*sin(\rotangle)}
            \pgfmathsetmacro{\newy}{\origy*cos(#4) + \zoffset*sin(#4)}
            \coordinate (#3) at (\newx, \newy);
        };
}

\newcommand{\createdocument}[4]{
    \coordinate (page-top-left-orig-#2) at (#1, 0);
    \coordinate (page-top-right-orig-#2) at ($(#1, 0) + (8, 0)$);
    \coordinate (page-bottom-left-orig-#2) at (#1, -11);
    \coordinate (page-bottom-right-orig-#2) at ($(#1, -11) + (8, 0)$);
    
    \calculateprojection{page-top-left-orig-#2}{page-top-left-#2}{#3}
    \calculateprojection{page-top-right-orig-#2}{page-top-right-#2}{#3}
    \calculateprojection{page-bottom-left-orig-#2}{page-bottom-left-#2}{#3}
    \calculateprojection{page-bottom-right-orig-#2}{page-bottom-right-#2}{#3}
    
    \draw[thick, fill=white] (page-top-left-#2) -- (page-top-right-#2) -- 
                             (page-bottom-right-#2) -- (page-bottom-left-#2) -- cycle;
    
    \draw[fill=black!20] ($(page-bottom-right-#2) + (0.1, -0.1)$) -- 
                          ($(page-bottom-left-#2) + (0.1, -0.1)$) -- 
                          (page-bottom-left-#2) -- 
                          (page-bottom-right-#2) -- cycle;
    \draw[fill=black!20] ($(page-bottom-right-#2) + (0.1, -0.1)$) -- 
                          ($(page-top-right-#2) + (0.1, -0.1)$) -- 
                          (page-top-right-#2) -- 
                          (page-bottom-right-#2) -- cycle;
    
    \coordinate (table-top-left-orig-#2) at ($(#1, 0) + (1, -2.5)$);
    \coordinate (table-top-right-orig-#2) at ($(#1, 0) + (7, -2.5)$);
    \coordinate (table-bottom-left-orig-#2) at ($(#1, 0) + (1, -8.5)$);
    \coordinate (table-bottom-right-orig-#2) at ($(#1, 0) + (7, -8.5)$);
    
    \calculateprojection{table-top-left-orig-#2}{table-top-left-#2}{#3}
    \calculateprojection{table-top-right-orig-#2}{table-top-right-#2}{#3}
    \calculateprojection{table-bottom-left-orig-#2}{table-bottom-left-#2}{#3}
    \calculateprojection{table-bottom-right-orig-#2}{table-bottom-right-#2}{#3}
    
    \draw[very thick] (table-top-left-#2) -- (table-top-right-#2) -- 
                      (table-bottom-right-#2) -- (table-bottom-left-#2) -- cycle;
    
    \coordinate (col1-top-orig-#2) at ($(table-top-left-orig-#2)!0.333!(table-top-right-orig-#2)$);
    \coordinate (col2-top-orig-#2) at ($(table-top-left-orig-#2)!0.667!(table-top-right-orig-#2)$);
    \coordinate (col1-bottom-orig-#2) at ($(table-bottom-left-orig-#2)!0.333!(table-bottom-right-orig-#2)$);
    \coordinate (col2-bottom-orig-#2) at ($(table-bottom-left-orig-#2)!0.667!(table-bottom-right-orig-#2)$);
    
    \calculateprojection{col1-top-orig-#2}{col1-top-#2}{#3}
    \calculateprojection{col2-top-orig-#2}{col2-top-#2}{#3}
    \calculateprojection{col1-bottom-orig-#2}{col1-bottom-#2}{#3}
    \calculateprojection{col2-bottom-orig-#2}{col2-bottom-#2}{#3}
    
    \draw[thick] (col1-top-#2) -- (col1-bottom-#2);
    \draw[thick] (col2-top-#2) -- (col2-bottom-#2);
    
    \coordinate (row1-left-orig-#2) at ($(table-top-left-orig-#2)!0.25!(table-bottom-left-orig-#2)$);
    \coordinate (row2-left-orig-#2) at ($(table-top-left-orig-#2)!0.5!(table-bottom-left-orig-#2)$);
    \coordinate (row3-left-orig-#2) at ($(table-top-left-orig-#2)!0.75!(table-bottom-left-orig-#2)$);
    \coordinate (row1-right-orig-#2) at ($(table-top-right-orig-#2)!0.25!(table-bottom-right-orig-#2)$);
    \coordinate (row2-right-orig-#2) at ($(table-top-right-orig-#2)!0.5!(table-bottom-right-orig-#2)$);
    \coordinate (row3-right-orig-#2) at ($(table-top-right-orig-#2)!0.75!(table-bottom-right-orig-#2)$);
    
    \calculateprojection{row1-left-orig-#2}{row1-left-#2}{#3}
    \calculateprojection{row2-left-orig-#2}{row2-left-#2}{#3}
    \calculateprojection{row3-left-orig-#2}{row3-left-#2}{#3}
    \calculateprojection{row1-right-orig-#2}{row1-right-#2}{#3}
    \calculateprojection{row2-right-orig-#2}{row2-right-#2}{#3}
    \calculateprojection{row3-right-orig-#2}{row3-right-#2}{#3}
    
    \draw[thick] (row1-left-#2) -- (row1-right-#2);
    \draw[thick] (row2-left-#2) -- (row2-right-#2);
    \draw[thick] (row3-left-#2) -- (row3-right-#2);
    
    \coordinate (header-bottom-left-orig-#2) at ($(table-top-left-orig-#2) + (0, -1.5)$);
    \coordinate (header-bottom-right-orig-#2) at ($(table-top-right-orig-#2) + (0, -1.5)$);
    \calculateprojection{header-bottom-left-orig-#2}{header-bottom-left-#2}{#3}
    \calculateprojection{header-bottom-right-orig-#2}{header-bottom-right-#2}{#3}
    \fill[gray!20] (table-top-left-#2) -- (table-top-right-#2) -- 
                   (header-bottom-right-#2) -- (header-bottom-left-#2) -- cycle;
    
    \coordinate (centroid-orig-#2) at ($(#1, 0) + (4, -5.5)$);
    \calculateprojection{centroid-orig-#2}{#2}{#3}
    
    \def\floatheight{1.5}
    
    \coordinate (node1-orig-#2) at ($(table-top-left-orig-#2) + (1, -2.25)$);
    \coordinate (node2-orig-#2) at ($(table-top-left-orig-#2) + (3, -2.25)$);
    \coordinate (node3-orig-#2) at ($(table-top-left-orig-#2) + (5, -2.25)$);
    
    \coordinate (node4-orig-#2) at ($(table-top-left-orig-#2) + (1, -3.75)$);
    \coordinate (node5-orig-#2) at ($(table-top-left-orig-#2) + (3, -3.75)$);
    \coordinate (node6-orig-#2) at ($(table-top-left-orig-#2) + (5, -3.75)$);
    
    \coordinate (node7-orig-#2) at ($(table-top-left-orig-#2) + (1, -5.25)$);
    \coordinate (node8-orig-#2) at ($(table-top-left-orig-#2) + (3, -5.25)$);
    \coordinate (node9-orig-#2) at ($(table-top-left-orig-#2) + (5, -5.25)$);
    
    \calculateprojectionz{node1-orig-#2}{\floatheight}{node1-#2}{#3}
    \calculateprojectionz{node2-orig-#2}{\floatheight}{node2-#2}{#3}
    \calculateprojectionz{node3-orig-#2}{\floatheight}{node3-#2}{#3}
    \calculateprojectionz{node4-orig-#2}{\floatheight}{node4-#2}{#3}
    \calculateprojectionz{node5-orig-#2}{\floatheight}{node5-#2}{#3}
    \calculateprojectionz{node6-orig-#2}{\floatheight}{node6-#2}{#3}
    \calculateprojectionz{node7-orig-#2}{\floatheight}{node7-#2}{#3}
    \calculateprojectionz{node8-orig-#2}{\floatheight}{node8-#2}{#3}
    \calculateprojectionz{node9-orig-#2}{\floatheight}{node9-#2}{#3}
    
    \calculateprojection{node1-orig-#2}{node1-base-#2}{#3}
    \calculateprojection{node2-orig-#2}{node2-base-#2}{#3}
    \calculateprojection{node3-orig-#2}{node3-base-#2}{#3}
    \calculateprojection{node4-orig-#2}{node4-base-#2}{#3}
    \calculateprojection{node5-orig-#2}{node5-base-#2}{#3}
    \calculateprojection{node6-orig-#2}{node6-base-#2}{#3}
    \calculateprojection{node7-orig-#2}{node7-base-#2}{#3}
    \calculateprojection{node8-orig-#2}{node8-base-#2}{#3}
    \calculateprojection{node9-orig-#2}{node9-base-#2}{#3}
    
    \foreach \i in {1,...,9} {
        \draw[#4!30, very thin, dashed] (node\i-base-#2) -- (node\i-#2);
    }
    
    
    
    \foreach \i in {1,...,9} {
        \foreach \j in {\i,...,9} {
            \ifnum\i<\j
                \pgfmathsetmacro{\randval}{rand}
                \pgfmathsetmacro{\threshold}{0.6} 
                \ifdim\randval pt>\threshold pt
                    \draw[#4!60, thick] (node\i-#2) -- (node\j-#2);
                \fi
            \fi
        }
    }
    
    \foreach \i in {1,...,9} {
        \fill[#4] (node\i-#2) circle (3pt);
        \draw[white, line width=0.5pt] (node\i-#2) circle (3pt);
    }
}

\createdocument{0}{doc1}{75}{red}
\createdocument{10}{doc2}{75}{blue}
\createdocument{24}{doc3}{45}{green}

\node[above=3cm of doc1, font=\Large\bfseries] (label-doc1) {Train Dataset A};
\node[above=3cm of doc2, font=\Large\bfseries] (label-doc2) {Train Dataset B};
\node[right=5cm of label-doc2, font=\Large\bfseries] {Finetune Dataset (test)};

\fill[black] (doc1) circle (2pt);
\fill[black] (doc2) circle (2pt);
\fill[black] (doc3) circle (2pt);

\newcommand{\placeGNN}[4]{
    \coordinate (gnn-center-#1) at ($(#1) + (0, #3)$);
    
    \node[draw=#2!70, fill=#2!10, thick, rounded corners=5pt, 
          minimum width=6.5cm, minimum height=4.75cm, align=center] 
          at (gnn-center-#1) (gnn-box-#1) {
        \textbf{\Large GNN Feature Extractor}\\[0.4cm]
        \large #4\\[0.3cm]
        \normalsize Node Embedding Layer
    };
    
\draw[->, thick, #2!60, line width=1.5pt] (#1) -- (gnn-box-#1.north) 
    node[pos=0.5, font=\small, #2!80, sloped, above] {Input Graph};
    
    \node[above=0.3cm of gnn-box-#1.south, text=black, font=\Large] {
        $h_i^{(k+1)} = \sigma\left(\sum_{j \in \mathcal{N}(i)} \theta^{(k)} h_j^{(k)}\right)$
    };

}

\newcommand{\placeStandaloneGNN}[4]{
    \node[draw=#3!70, fill=#3!10, thick, rounded corners=5pt, 
          minimum width=9cm, minimum height=4.75cm, align=center] 
          at #2 (gnn-box-#1) {
        \textbf{\Large GNN Feature Extractor}\\[0.4cm]
        \large #4\\[0.3cm]
        \normalsize Node Embedding Layer
    };
    
    \node[above=0.3cm of gnn-box-#1.south, font=\Large, text=black] {
        $        h_i^{(k+1)} = \sigma\left(\sum_{j \in \mathcal{N}(i)} \frac{1}{N}\sum_{d_n \in D} \theta^{(k, d_n)} h_j^{(k)}\right)$
    };
}

\placeGNN{doc1}{red}{-6}{Graph Neural Network}
\placeGNN{doc2}{blue}{-6}{Graph Neural Network}

\node[circle, draw=black, thick, fill=white, minimum size=1cm, font=\Large\bfseries] 
    at ($(gnn-box-doc1)!0.5!(gnn-box-doc2)$) {$+$};

\placeStandaloneGNN{merged}{($(gnn-box-doc2) + (11, 0)$)}{green}{Merged Message Passing}

\draw[->, thick, black!60, line width=1.5pt] (gnn-box-doc2.east) -- (gnn-box-merged.west)
    node[midway, above, font=\small, black!80] {Aggregate};
    
\end{tikzpicture}

%% file: data/bigtable/table_ws_summary.tex
\begin{table}[t]
\centering
\resizebox{\textwidth}{!}{%
\begin{tabular}{lc}
\toprule
\multicolumn{2}{c}{\textbf{mAP}} \\
\toprule
Mean mAP (C / D)   & .313 / \textbf{.363 }\\
\hline
\multicolumn{2}{c}{\textbf{Recall}} \\
\hline
R@1 (C/D)  & .256/\textbf{.300} \\
R@5 (C/D)  & .388/\textbf{.451} \\
R@10(C/D)  & .447/\textbf{.516} \\
\hline
\end{tabular}
\hspace{2em}
\begin{tabular}{lccc}
\toprule
Stat & ID & OOD-L & OOD-U \\
\toprule
Abs. gain & .005 & .119 & .034 \\
Gain (\%)  & 11.58\% & 36.68\% & 104.02\% \\
Std gain  & .109 & .056 & .063 \\
Min gain  & -.228 & .050 & -.181 \\
Max gain  & .146 & .222 & .172 \\
\hline
\end{tabular}
\hspace{2em}
\begin{tabular}{lcc}
\toprule
Encoder & Regime & mAP ($\pm$ std) \\
\toprule
\multirow{2}{*}{GRU} & C & .356 $\pm$ .294 \\
 & D & \textbf{.399} $\pm$ .299 \\
 \midrule
\multirow{2}{*}{LSTM} & C & .358 $\pm$ .293 \\
 & D & \textbf{.412} $\pm$ .301 \\
 \midrule
\multirow{2}{*}{Transformer} & C & .184 $\pm$ .205 \\
 & D & \textbf{.246 }$\pm$ .238 \\
\hline
\end{tabular}%
}
\caption{Centralized (C) vs. distributed (D) pre-training: average recall across word spotting datasets, including ID, OOD-L, and OOD-U scenarios.}
\label{tab:comparison}
\end{table}